\pdfoutput=1

\documentclass[11pt]{article}

\usepackage[]{acl}

\usepackage{times}
\usepackage{latexsym}

\usepackage[T1]{fontenc}

\usepackage[utf8]{inputenc}

\usepackage{microtype}

\usepackage{inconsolata}

\usepackage{microtype}
\usepackage{graphicx}
\usepackage{multirow}
\usepackage{subfigure}
\usepackage{amsmath, amssymb, mathtools}
\usepackage[export]{adjustbox}
\usepackage{hyperref}

\title{Semi-Supervised Knowledge-Grounded Pre-training for Task-Oriented Dialog Systems}

\author{Weihao Zeng$^{1*}$, Keqing He$^{2*}$, Zechen Wang$^{1*}$, Dayuan Fu$^{1}$, Guanting Dong$^{1}$ \\ {\bf Ruotong Geng$^{1}$,}  {\bf Pei Wang$^{1}$,} {\bf Jingang Wang$^{2}$,} {\bf Chaobo Sun$^{2}$,} {\bf Wei Wu$^{2}$,} {\bf Weiran Xu$^{1}$}\thanks{\ \ The first three authors contribute equally. Weiran Xu is the corresponding author.}\\
  $^1$Beijing University of Posts and Telecommunications, Beijing, China\\
$^{2}$Meituan, Beijing, China\\
  \texttt{\{zengwh,zechen\_wang,fdy,dongguanting,ruotonggeng,wangpei,xuweiran\}@bupt.edu.cn}\\
  \texttt{\{hekeqing,wangjingang,sunchaobo,wuwei\}@meituan.com}
  }

\begin{document}
\maketitle
\begin{abstract}
Recent advances in neural approaches greatly improve task-oriented dialogue (TOD) systems which assist users to accomplish their goals. However, such systems rely on costly manually labeled dialogs which are not available in practical scenarios. In this paper, we present our models for Track 2 of the SereTOD 2022 challenge, which is the first challenge of building semi-supervised and reinforced TOD systems on a large-scale real-world Chinese TOD dataset MobileCS. We build a knowledge-grounded dialog model to formulate dialog history and local KB as input and predict the system response. And we perform semi-supervised pre-training both on the labeled and unlabeled data. Our system achieves the first place both in the automatic evaluation and human interaction, especially with higher BLEU (+7.64) and Success (+13.6\%) than the second place.\footnote{Our code, models and other related resources are publicly
available at \href{https://github.com/Zeng-WH/S2KG}{https://github.com/Zeng-WH/S2KG}.}
\end{abstract}

\section{Introduction}

Task-oriented dialogue (TOD) systems assist users to accomplish their goals like booking a ticket and make an effect on everyone's lives with recent advances in neural approaches \cite{gao-etal-2018-neural-approaches}. A typical TOD system consists of three sub-modules: (1) natural language understanding (NLU) for recognizing the user's intent and slots \cite{Goo2018SlotGatedMF,Qin2019ASF,He2020LearningTT,Xu2020ADG,He2020ContrastiveZL}; (2) dialog management (DM) for tracking dialog states \cite{WuTradeDST2019,Gao2019DialogST} and deciding which system action to take \cite{Peng2018DeepDynaQ,Liu2021ScheduledDP}; (3) natural language generation (NLG) for generating dialogue response corresponding to the predicted system action \cite{peng-etal-2020-shot}. Traditional modular methods \cite{Goo2018SlotGatedMF,WuTradeDST2019,peng-etal-2020-shot} and recent end-to-end modeling methods \cite{Peng2021SoloistBT,Su2022MultiTaskPF,Liu2022RevisitingMG} achieve decent performance in several or all modules. However, such systems rely on costly manually labeled dialogs which are not available in practical scenarios. It's valuable to explore semi-supervised learning (SSL) \cite{Zhu2005SemiSupervisedLL} for TOD, which aims to leverage both labeled and unlabeled data. 

To facilitate relevant research, SereTOD 2022 Workshop \footnote{http://seretod.org/} proposes the first challenge of building semi-supervised and reinforced TOD systems by releasing a large-scale Chinese TOD dataset MobileCS from real-world dialog transcripts between real users and customer-service staffs from China Mobile. MobileCS contains 10,000 labeled dialogs and 90,000 unlabeled dialogs. There are two tracks: (1) Information extraction (Track 1) aims to extract entities together with their slot values. (2) Task-oriented dialog system (Track 2) aims to build a complete TOD system, including predicting the user intent, querying the local KB, and generating appropriate system intent and response according to the given dialog history. The core challenge is how to combine a small labeled dataset and a large unlabeled dataset.

\begin{figure*}[t]
\centering
\resizebox{0.9\textwidth}{!}{
\includegraphics[scale=0.65]{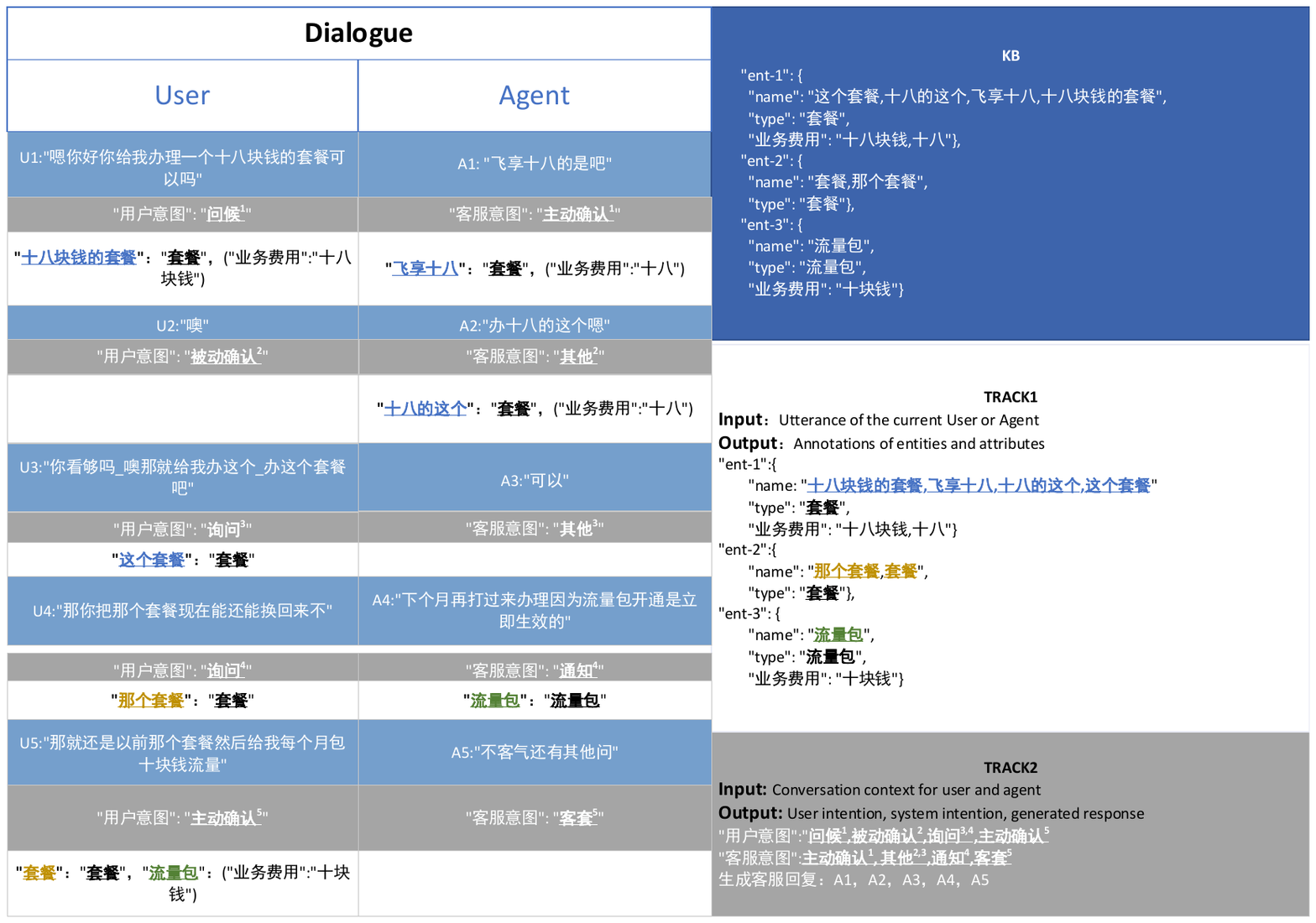}}
\caption{An example from MobileCS.}
\label{fig:data_example}
\end{figure*}

In this paper, we present our system for Track 2 of the SereTOD 2022 challenge. The main intuition behind our system comes from semi-supervised knowledge-grounded pre-training on both labeled and unlabeled datasets. We divide Track 2 into two task groups, classification (user intent and system intent) and generation (system response). For the classification tasks, we employ Roberta-large \footnote{https://huggingface.co/hfl/chinese-roberta-wwm-ext-large} and build two separate classification models. We also perform continual pre-training on all the dialog data. For the generation task, we build a knowledge-grounded dialog model, which is the key point of this paper. Specifically, we firstly use pre-trained language models (e.g. T5 \footnote{https://github.com/ZhuiyiTechnology/t5-pegasus} and UFA \cite{He2022UnifiedKP}) as our backbone. Then, we take the dialog history and serialized local KB \footnote{A local KB for a dialog could be viewed as being composed of the relevant snapshots from the global KB. Please see more details in \citet{Ou2022ACO}.} as input and output system response. Here, we simply concatenate each key-value pair in the local KB as \emph{key: value} to build a string input. We only use response generation as the learning objective. For the labeled dataset, we use the golden KB annotations as our input. For the unlabeled dataset, we obtain the predicted KB results using our model in Track 1. Finally, we mix up all the data to train a knowledge-grounded dialog model.

We summarize the main contributions of our system S2KG (\textbf{S}emi \textbf{S}upervised \textbf{K}nowledge-\textbf{G}rounded pre-training) as follows:

\begin{itemize}
    \item We build a knowledge-grounded dialog model to formulate dialog history and local KB as input and predict the system response.
    \item We perform semi-supervised pre-training both on the labeled and unlabeled data.
\end{itemize}
Our system achieves the first place both in the automatic evaluation and human interaction, especially with higher BLEU (+7.64) and Success (+13.6\%) than the second place.

\section{Task Description}
\begin{table}[]
\centering
\begin{tabular}{lcc}
\hline
\textbf{Metric}      & \multicolumn{1}{l}{\textbf{labeled}} & \multicolumn{1}{l}{\textbf{unlabeled}} \\ \hline
Dialogs              & 8,975                                & 87,933                                 \\
Turns                & 100,139                              & 972,573                                \\
Tokens               & 3,991,197                            & 39,491,883                             \\
Avg.turns per dialog & 11.16                                & 11.06                                  \\
Avg.tokens per turn  & 39.86                                & 40.61                                  \\
Slots                & 26                                   & -                                      \\
Values               & 14,623                               & -                                      \\ \hline
\end{tabular}
\caption{Training dataset statistics of MobileCS. The challenge also provides another 1,000 labeled dialogs as evaluation data (dev set).}
\label{tab:data_stat}
\end{table}

MobileCS is a large Chinese TOD dataset collected from real-world dialog transcripts between real users and customer-service staffs. Different from the simulated MultiWOZ dataset \cite{Budzianowski2018MultiWOZA}, it consists of real-life data and large unlabeled dialogs. Specifically, MobileCS contains 10,000 labeled dialogs and 90,000 unlabeled dialogs. The full data statistics are shown in Table \ref{tab:data_stat}. The challenge has two tracks. Track 1 (information extraction) aims to extract entities and attributes to build a local knowledge base (KB). And Track 2 uses the KB and raw dialogs to train a complete TOD system. We provide a real annotated dialog in Figure \ref{fig:data_example}. In this paper, we focus on Track 2. Here, we elaborate on the task details. Track 2 for the TOD system is, for each dialog turn, given the dialog history, the user utterance and the local KB, to predict the user intent, query the local KB and generate appropriate system intent and response according to the queried information. For every labeled dialog, the annotations consist of user intents, system intents and a local KB. The local KB is obtained by collecting the entities and triples annotated for Track 1. For unlabeled dialogs, there are no such annotations. 

\begin{figure}[t]
\centering
\resizebox{0.5\textwidth}{!}{
\includegraphics[scale=0.5]{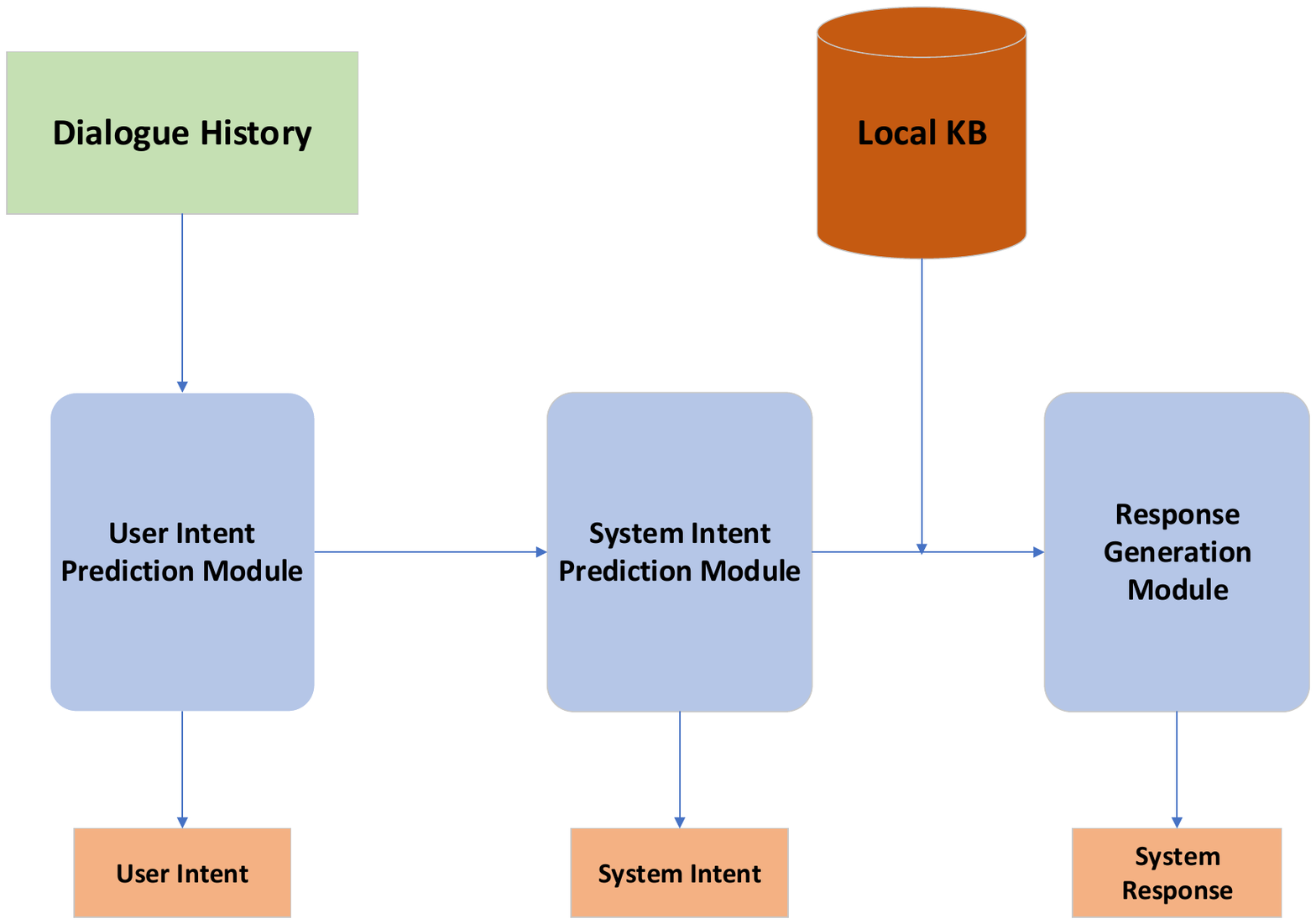}}
\caption{Overall architecture of our knowledge-grounded task-oriented dialogue system.}
\label{fig:overall}
\end{figure}

To measure the performance of TOD systems, both automatic evaluation and human evaluation will be conducted. For automatic evaluation, metrics include Precision/Recall/F1 score, Success rate and BLEU \cite{Papineni2002BleuAM} score. P/R/F1 are calculated for both predicted user intents and system intents. Success rate is the percentage of generated dialogs that achieve user goals. BLEU score evaluates the fluency of generated responses\footnote{The challenge adopts BLEU-4.}. For human evaluation for different TOD systems, real users will interact with those systems according to randomly given goals. For each dialog, the user will score the system on a 5-point scale (1-5) by the following 3 metrics. 5 denotes the best and 1 denotes the worst, respectively.
\begin{itemize}
    \item \textbf{Success}. This metric measures if the system successfully completes the user goal by interacting with the user;
    \item \textbf{Coherency}. This metric measures whether the system's response is logically coherent with the dialogue context;
    \item \textbf{Fluency}. The metric measures the fluency of the system's response.
\end{itemize}
The average score from automatic evaluation and human evaluation is the main ranking basis on the leaderboard. 

\section{Methodology}
\subsection{Overall Architecture}
Figure \ref{fig:overall} shows the overall system architecture for Track 2. Track 2 contains three tasks: user intent, system intent, and system response. For the user intent and system intent tasks, we use Roberta-large and build two separate classification models. For the system response task, we build a knowledge-grounded dialog model and perform semi-supervised pre-training both on the labeled and unlabeled data.

\begin{figure}[t]
\centering
\resizebox{0.5\textwidth}{!}{
\includegraphics[scale=0.5]{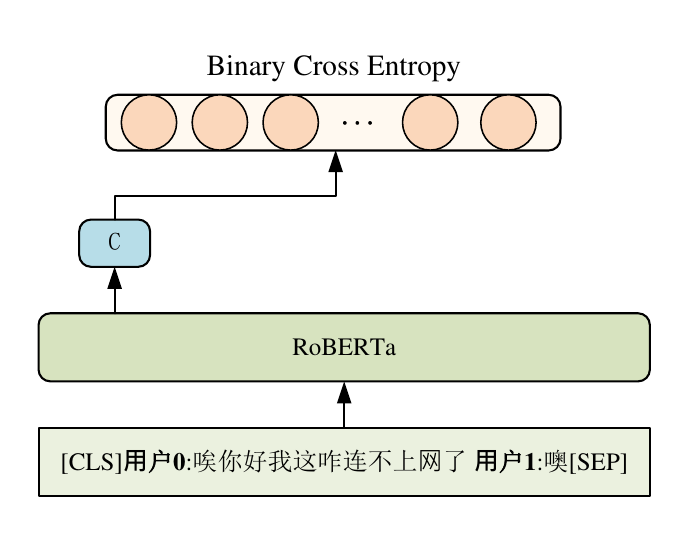}}
\caption{The architecture of the classification models.}
\label{fig:nlu_model}
\end{figure}

\subsection{Subtask 1: Classification}
Given a dialog history, the user intent and system intent tasks aim to predict the user intent or system intent(act) respectively. Considering both the tasks are multi-label, we formulate the tasks as multi-label text classification questions. As Figure \ref{fig:nlu_model} displays, we adopt Roberta as our backbone and use the dialog history as input. For the user intent task, we concatenate two user utterances as input. We find too many turns bring no further improvements and introducing system responses has a side effect. We suppose the gap between training and prediction affects the model performance. For the system intent task, we concatenate three user utterances as input.\footnote{The system intent task also requires intent arguments. We use heuristic rules based on the local KB to match the entities.} Then, we use the hidden state of the [CLS] token to predict the results. Binary cross entropy is the learning objective. Section \ref{nlu} proves classification models outperform GPT-based end-to-end models. Besides, we also introduce some augmentation strategies as follows:
\begin{itemize}
    \item \textbf{Continual Pre-training}. We pre-train Roberta on the labeled and unlabeled dialogs using MLM objective like BERT \cite{devlin-etal-2019-bert}. We pre-train 20 epochs using a learning rate of 5e-4 and 15\% mask rate. MLM continual pre-training brings large improvements of 1.68\% on User Intent F1 and 1.86\% on System Intent F1.
    \item \textbf{Class-wise Threshold}. We adaptively select the best threshold for each intent type based on the performance on the dev set. This strategy brings improvements of 1.23\% on User Intent F1 and 1.48\% on System Intent F1.
    \item \textbf{Adversarial Training}. We adopt FGM \cite{Goodfellow2015ExplainingAH} as our adversarial training strategy. This strategy brings improvements of 0.64\% on User Intent F1 and 0.46\% on System Intent F1. 
\end{itemize}

\begin{figure}[t]
\centering
\resizebox{0.5\textwidth}{!}{
\includegraphics[scale=0.5]{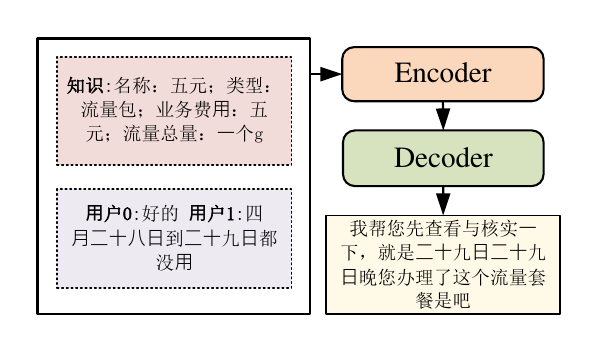}}
\caption{The architecture of the generation model.}
\label{fig:nlg_model}
\end{figure} 

\begin{table*}[t]
\centering
\resizebox{0.85\textwidth}{!}{
\begin{tabular}{l|ccccc}
\hline
\multirow{2}{*}{\textbf{Team ID}}
  & \multicolumn{5}{c}{\textbf{Automatic Evaluation}}                                                           \\
       & \textbf{User Intent F1} & \textbf{System Intent F1} & \textbf{BLEU}  & \textbf{Success} & \textbf{Combined} \\ \hline
\textbf{Team-11 (Ours)} & \textbf{0.728}                   & \textbf{0.595}            & \textbf{14.430} & \textbf{0.780}    & \textbf{2.392}    \\
Team-5                 & 0.714                   & 0.589                     & 6.790           & 0.432            & 1.871             \\
Team-13                & 0.706                   & 0.587                     & 5.526          & 0.251            & 1.655             \\
Team-10                & 0.664                   & 0.504                     & 3.629          & 0.217            & 1.458             \\
Team-8                 & 0.699                   & 0.550                      & 6.440           & 0.644            & 2.022             \\
\hline
official baseline               & 0.644                   & 0.394                     & 4.170          & 0.315            & 1.436             \\ \hline
\end{tabular}
}
\caption{Final automatic results on the test set of the top 5 teams released by the officials. User Intent F1 denotes the performance of classifying the input user query and System Intent F1 denotes the predicted system acts. Success rate is the percentage of generated dialogs that achieve user goals. Combined score is the overall result which is calculated as follows: Combined score = User intent F1 + System intent F1 + Success + BLEU/50.}
\label{tab:automatic}
\end{table*}

\subsection{Subtask 2: Generation}
For the generation task, we build a knowledge-grounded dialog model, S2KG in Figure \ref{fig:nlg_model}. Specifically, we firstly use pre-trained language models (e.g. T5 \footnote{https://github.com/ZhuiyiTechnology/t5-pegasus} and UFA \cite{He2022UnifiedKP}) as our backbone. Then, we take the dialog history and serialized local KB as input and output system response. Here, we simply concatenate each key-value pair in the local KB as \emph{key: value} to build a string input. We only use response generation as the learning objective. We find KB grounding has a large improvement over baselines (see Section \ref{nlg}).

The SereTOD challenge gives a large-scale (90,000) unlabeled dataset that doesn't contain KB annotations and a relatively small (10,000) labeled dataset. So we perform semi-supervised pre-training to utilize all these dialogs. For the labeled dataset, we use the golden KB annotations as our input. For the unlabeled dataset, we obtain the predicted KB results using our model in Track 1. We implement our system of Track 1 mainly based on the official baseline \cite{Liu2022InformationEA}. Finally, we mix up all the data to train a knowledge-grounded dialog model. We find only using unsupervised pre-training gets an improvement of 1.91 BLEU, but drops by 14.6 on Success, because raw response generation pre-training makes the model memorize similar dialogs but predict unfaithful responses without grounding ability. Therefore, it's necessary to obtain pseudo KB annotations to perform pre-training. Note that the performance of the Track 1 system is relatively poor so we argue the quality of pseudo KB makes no significant effect on the final results. We leave more discussion to future work.

\section{Experiment}

\subsection{Setup}
We train our models on the training set and report our results on the dev set. The final leaderboard results are evaluated on the test set. Since the test set is not released until the end of the challenge, we perform ablation studies only on the dev set. We conduct our experiments using Huggingface\footnote{https://huggingface.co/} and computation platform JiuTian\footnote{https://jiutian.10086.cn/edu/\#/home}.

\begin{table}[]
\centering
\resizebox{0.48\textwidth}{!}{%
\begin{tabular}{l|cc}
\hline
\multicolumn{1}{c|}{\textbf{Methods}}        & \textbf{User Intent F1} & \textbf{System Intent F1} \\ \hline
GPT-2 (baseline)                              & 0.6488         & 0.4012           \\ \hline
Roberta                    & 0.7448         & 0.5158           \\
Roberta+FGM                & 0.7512         & 0.5204           \\
Roberta+FGM+Threshold     & 0.7635         & 0.5352           \\
Roberta+FGM+Threshold+MLM & \textbf{0.7803}         & \textbf{0.5538}           \\ \hline
\end{tabular}
}
\caption{Comparison of different user intent and system intent models on the dev set.}
\label{tab:nlu}
\end{table}

\subsection{Main Results}
Table \ref{tab:automatic} shows the final automatic results on the test set of the top 5 teams\footnote{See all the results in the \href{https://docs.google.com/spreadsheets/d/1w28AKkG6Wjmuo15QlRlRyrnv859MT1ry0CHV8tFxY9o/edit?usp=sharing}{official leaderboard}.}. Our system (Team 11) achieves the state-of-the-art on all metrics, especially for generation task, demonstrating the effectiveness of our proposed S2KG. Specifically, our method outperforms the second place (Team 5) by 1.4\% on User Intent F1 and 0.6\% on System Intent F1. The improvements mainly come from better pre-trained LM, continual pre-training, class-wise threshold, and adversarial training. We will dive into details in Section \ref{nlu}. For generation metrics, our S2KG model significantly outperforms the second place with a large margin of 7.640 on BLEU and 34.8\% on Success. The improvements are mainly attributed to knowledge-grounded dialog model and semi-supervised pre-training, which are the key points of this paper. We leave the discussion to Section \ref{nlg}.

\begin{table*}[t]
\centering
\resizebox{0.80\textwidth}{!}{
\begin{tabular}{l|cccc|c}
\hline
\multirow{2}{*}{\textbf{Team ID}}  & \multicolumn{4}{c|}{\textbf{Human Evaluation}}                              & \multirow{2}{*}{\textbf{Final Score}} \\ 
       & \textbf{Fluency} & \textbf{Coherency} & \textbf{Success} & \textbf{Average} &                                       \\ \hline
\textbf{Team-11 (Ours)} & \textbf{4.23}             & \textbf{3.73}      & \textbf{3.47}    & \textbf{3.81}    & \textbf{3.10 }                                 \\
Team-5                 & 4.06             & 3.14               & 3.40              & 3.53             & 2.70                                  \\
Team-13                & 3.55             & 3.03               & 2.77             & 3.12             & 2.39                                  \\
Team-10                & 3.20              & 2.98               & 3.11             & 3.10              & 2.28                                  \\
Team-8                 & 2.39             & 2.29               & 2.03             & 2.24             & 2.13                                  \\ \hline
\end{tabular}
}
\caption{Final human evaluation results on the test set of the top 5 teams released by the officials. Final score is the average of Combined score from automatic evaluation and  averaged human evaluation score. It's the main ranking basis on the Track 2 leaderboard.}
\label{tab:human}
\end{table*}

\begin{table}[]
\centering
\resizebox{0.45\textwidth}{!}{
\begin{tabular}{l|ll}
\hline
\multicolumn{1}{c|}{\textbf{Methods}} & \multicolumn{1}{c}{\textbf{BLEU}} & \multicolumn{1}{c}{\textbf{Success}} \\ \hline
GPT2-FT (baseline)                              & 4.39                              & 0.344                                \\
GPT2-KGFT                             & 7.48                              & 0.692                                \\ \hline
T5-KGFT                       & 11.32                             & 0.741                                \\
T5-Unsup-KGFT                 & 13.23                             & 0.595                                \\
T5-Semi               & 14.39                             & 0.767                                \\
T5-Semi-KGFT               & 12.30                             &   0.761                              \\ \hline
UFA-Semi                 & 14.51                             & 0.789                                \\ \hline
\end{tabular}
}
\caption{Comparison of different system response generation models on the dev set.}
\label{tab:nlg}
\end{table}

\subsection{Classification}
\label{nlu}
To verify the effect of our proposed models, we perform an ablation study of different user intent and system intent models on the dev set in Table \ref{tab:nlu}. GPT-2 is the official baseline \cite{Liu2022RevisitingMG} which is an end-to-end generative model based on Chinese GPT-2\footnote{https://huggingface.co/uer/gpt2-chinese-cluecorpussmall}. For pre-trained language models, we find Roberta-based classification models get better performance with improvements of 9.60\% on User Intent F1 and 11.46\% on System Intent F1. Based on Roberta, we also introduce some training or inference strategies, including adversarial training FGM, class-wise threshold, and MLM continual pre-training. All the strategies show advantages. MLM continual pre-training brings the largest improvements of 1.68\% on User Intent F1 and 1.86\% on System Intent F1, demonstrating the effectiveness of pre-training on domain corpus. Other strategies also get 0.5-1\% improvements.

\subsection{Generation}
\label{nlg}

Table \ref{tab:nlg} displays the ablation study of our S2KG system for the response generation task. We analyze the results from the following perspectives.

\textbf{Knowledge Grounding} GPT2-FT (finetune) denotes the official baseline. GPT2-KGFT is our proposed knowledge grounding finetuning method which uses the serialized KB as knowledge. The first two lines in Table \ref{tab:nlg} show GPT2-KGFT significantly outperforms GPT2-FT by 3.09 BLEU and 34.8\% Success, demonstrating the effectiveness of knowledge grounding based on local KB. We also find knowledge grounding improves the factual consistency of generated responses. We give examples in Section \ref{Success}. 

\textbf{Semi-Supervised Pre-training} The SereTOD challenge gives a large-scale unlabeled dataset that doesn't contain KB annotations. So we perform different pre-training settings to utilize these unlabeled dialogs. T5-KGFT is our proposed knowledge grounding model which replaces GPT2 with T5. Based on T5-KGFT, T5-Unsup-KGFT first performs an unsupervised response generation pre-training without KB input and then adopts knowledge grounding finetuning. Results show unsupervised pre-training gets an improvement of 1.91 BLEU, but drops by 14.6 on Success. We argue it's because raw response generation pre-training makes the model memorize similar dialogs but predict unfaithful responses without grounding ability. T5-Semi replaces unsupervised pre-training with semi-supervised pre-training which uses Track 1 system to generate the pseudo local KB for these unlabeled dialogs. T5-Semi outperforms T5-Unsup-KGFT by 1.16 BLEU and 17.2\% Success, demonstrating the effectiveness of semi-supervised pre-training. We also find continual knowledge grounding finetuning on labeled data (T5-Semi-KGFT) can't bring further improvements upon T5-Semi because of knowledge forgetting.

\begin{figure*}[t]
\centering
\resizebox{0.9\textwidth}{!}{
\includegraphics[scale=0.65]{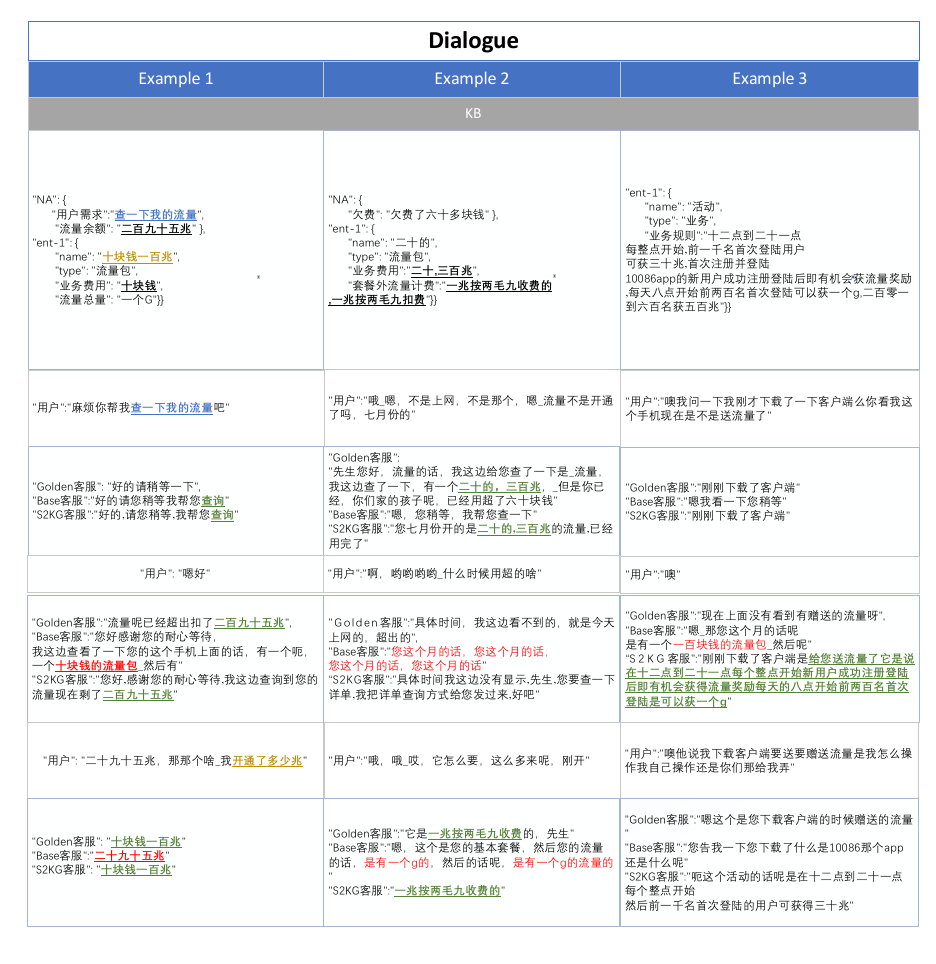}}
\vspace{-0.7cm}
\caption{Case study for three examples from Baseline and S2KG system. We present the local KB, the user utterances, golden response, baseline predictions, and prediction of S2KG system.}
\label{fig:success}
\end{figure*}
 
\textbf{Pre-trained Language Model} We also compare different PLMs. We find that T5 consistently achieves better results than GPT-2. Besides, we experiment with a large PLM specified for customer service, UFA-large \cite{He2022UnifiedKP}, which has 1.2B parameters compared to 220M T5 and 117M GPT-2. UFA-large further outperforms T5 by 0.12 BLEU and 2.2\% Success.\footnote{Considering the inference efficiency and hardware limit, we submit our final results on T5.}

\subsection{Human Evaluation}
SereTOD performs human evaluation for different TOD systems, where real users interact with those systems according to randomly given goals. Table \ref{tab:human} shows the results of human evaluation and final scores. Our system also achieves state-of-the-art on all the metrics. Specifically, our method outperforms the second place (Team 5) by 0.17 on Fluency, 0.59 on Coherency, and 0.07 on Success.

\section{Analysis}

 \begin{figure*}[t]
\centering
\resizebox{0.9\textwidth}{!}{
\includegraphics[scale=0.65]{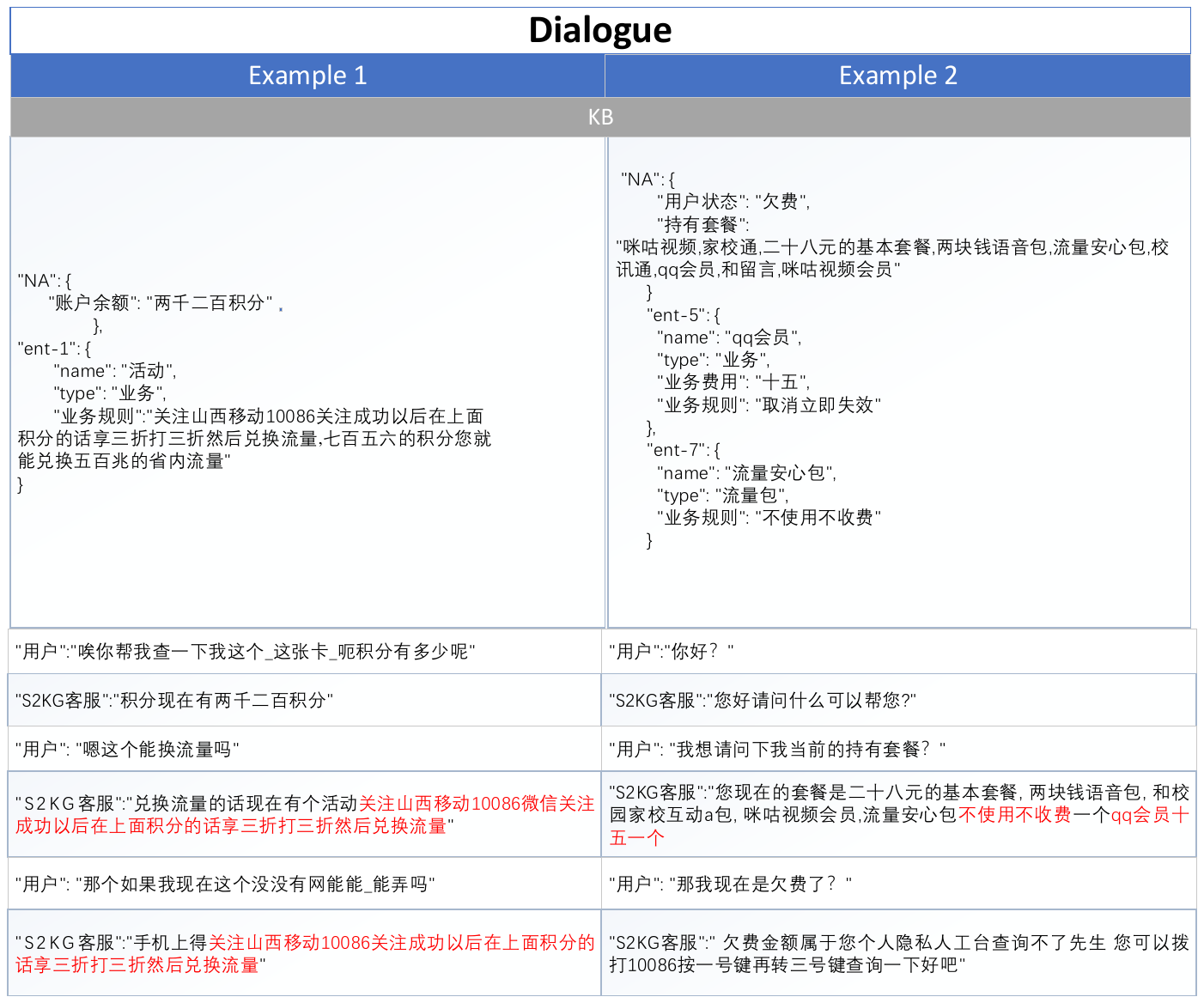}}
\vspace{-0.7cm}
\caption{Case study for two examples from S2KG system. We present the local KB, the user utterances and prediction of S2KG system.}
\label{fig:challenge}
\end{figure*} 

\subsection{Case Study}

Figure \ref{fig:success} shows three examples from the baseline model and the S2KG model, respectively, proving the advantages of S2KG model from the three dimensions of Success, Fluency, and Coherency.

\label{Success}
\noindent\textbf{Success} In example one, the local KB includes the user's mobile package balance and information about the data package plan currently held by the user. The user's utterance is "Could you please check my data package for me?", which means the user asks the system to query the mobile package balance. The baseline system misidentified the user's intent and mistakenly believed that the user was querying the information of the data package plan, so it retrieved the wrong knowledge "ten yuan data package plan", thereby generating a reply wrongly. The S2KG model correctly identified the user's request, retrieved the correct result based on local KB, and successfully answered that the current mobile package balance was 295M in the reply. It proves that knowledge-grounded semi-supervised pre-training can greatly improve the accuracy of knowledge selection.

\noindent\textbf{Fluency} In example two, the user's second round of utterance is intended to query the date when the data package cap is exceeded. Since there is no corresponding information in the current local KB, the system cannot retrieve the knowledge. In this scenario, the baseline system repeated meaninglessly and failed to generate fluent responses. Due to the large-scale pre-training, the S2KG model can explain the situation to the user, guide the user correctly, and provide the user with a reasonable solution.

\noindent\textbf{Consistency} In example three, the user's dialogue history is mainly related to the data package given by the activity, while the baseline model mainly answers the user's current data package in the second round of replies, which cannot be consistent with the dialogue history. The S2KG model has good modeling of the dialogue history through pre-training, so it can explain the activity rules to the user in detail, to meet the user's intent, and the generated replies are consistent with the dialogue history.

\subsection{Challenge}

Although the S2KG model has achieved SOTA in the three dimensions of Success, Fluency and Coherency, there are still issues unresolved as showed in Figure \ref{fig:challenge}: (1) \textbf{Response Diversity}: As shown in example 1, the user wants to figure out the rules for the user points redemption activity. Although the system provides the retrieved plan rules, the user still cannot understand these rules. So the user asks a question again, then the system repeats the business rules in the KB, resulting in a decrease in the diversity of the response and the user's engagement. (2) \textbf{Knowledge Redundancy}: As shown in example 2, the user asks the system about the data packages the user hold, the S2KG model not only provides the name of the packages in the reply, but also retrieves other information associated with the packages, such as fees, etc. As a result, there is knowledge redundancy in the system's reply, which is not conducive to the user's ability to grasp the key points.

\section{Conclusion}
In this paper, we present our models for Track 2 of the SereTOD 2022 challenge aiming to build semi-supervised and reinforced TOD systems. We divide Track 2 into two task groups, classification (user intent and system intent) and generation (system response). For the classification tasks, we employ Roberta-large and build two separate classification models. We also perform continual pre-training, class-wise threshold, and adversarial training strategies. For the generation task, we build a knowledge-grounded dialog model S2KG and perform semi-supervised pre-training both on labeled data and unlabeled data. Our system achieves first place both in the automatic evaluation and human interaction. We also discuss the advantages and challenges of our system to provide a guideline for future work.

\section*{Acknowledgements}
We thank all anonymous reviewers for their helpful comments and suggestions. We are also grateful to the track organizers for their valuable work. This work was partially supported by National Key R\&D Program of China No. 2019YFF0303300 and Subject II No. 2019YFF0303302, DOCOMO Beijing Communications Laboratories Co., Ltd, MoE-CMCC "Artifical Intelligence" Project No. MCM20190701.


\bibliography{anthology,custom}
\bibliographystyle{acl_natbib}


\end{document}